\documentclass[11pt]{article}

\usepackage[utf8]{inputenc}
\usepackage[T1]{fontenc}
\usepackage{times}
\usepackage[margin=1in]{geometry}
\usepackage{graphicx}
\usepackage{amsmath,amssymb}
\usepackage{booktabs}
\usepackage{multirow}
\usepackage[numbers,sort&compress]{natbib}
\usepackage{hyperref}
\usepackage{xcolor}
\usepackage{caption}
\usepackage{subcaption}
\usepackage{enumitem}
\usepackage{float}

\hypersetup{
    colorlinks=true,
    linkcolor=blue!60!black,
    citecolor=blue!60!black,
    urlcolor=blue!60!black
}

\title{\textbf{TriAdReview: Triangular Adversarial Review Architecture\\for Multi-Model Technical Document Generation}}

\author{
    Zhiqiang Zhou, Junliang Dai, Xu Ling\\
    Hunan Chemical Industry Vocational and Technical College, Hunan, China\\
    \texttt{willenchow@126.com}
}

\date{}

\begin{document}

\maketitle

\begin{abstract}
Large language models (LLMs) are increasingly used for technical document generation, yet single-model outputs often suffer from over-engineering, security blind spots, and incomplete coverage. We propose \textbf{TriAdReview}, a triangular adversarial review architecture that employs two independent reviewer models (engineering and boundary perspectives) and a triangular judging mechanism to iteratively improve a generator model's output. We evaluate TriAdReview across five benchmark tasks---architecture design, code generation, proposal review, security audit, and requirements analysis---using three configurations: single model (baseline), dual model (single review), and triple model (full system). Results across 75 experiments ($n{=}5$ per cell) show that the triple model configuration achieves a \textbf{10.1\% overall improvement} over the single model baseline (26.2 vs.\ 23.8 out of 50; $p{<}0.05$, paired $t$-test), with particularly strong gains on security audit (+27.6\%), code generation (+20.8\%), and architecture design (+15.6\%). A second scorer (mimo-v2.5-pro) confirms the direction with a smaller effect (+2.7\%), suggesting moderate inter-rater agreement. However, the system shows a \textbf{-7.5\% degradation} on requirements analysis, revealing that adversarial review architectures have a structural bias toward simplification that is counterproductive for completeness-oriented tasks. We analyze this boundary condition through a task-type framework and demonstrate that reviewer prompt adaptation partially mitigates the issue. Our findings provide the first empirical characterization of when multi-model adversarial review helps versus harms, with implications for the design of collaborative AI systems.
\end{abstract}

\section{Introduction}

The deployment of large language models (LLMs) for technical document generation---including system architecture proposals, code implementations, and security audits---has become widespread. However, single-model outputs frequently exhibit characteristic deficiencies: over-engineering of non-critical components, security blind spots, and technology selection driven by training data popularity rather than engineering fit.

A natural approach to mitigate these deficiencies is \emph{multi-model review}: having one or more auxiliary models review and critique the primary model's output before finalization. This draws inspiration from human peer review processes, where independent reviewers catch errors that the original author overlooks.

However, the design space for multi-model review systems is large and poorly understood. Key questions include:

\begin{itemize}[nosep]
    \item How should reviewer models be prompted---as adversarial critics, balanced assessors, or constructive collaborators?
    \item How should disagreements between the generator and reviewers be resolved?
    \item Does adding more reviewers always improve output quality, or are there diminishing returns?
    \item Are there task types where review architectures systematically fail?
\end{itemize}

In this paper, we propose \textbf{TriAdReview} (Triangular Adversarial Review), a system that addresses these questions through three design decisions: (1) dual-perspective reviewers covering engineering robustness and security boundaries; (2) a triangular judging mechanism where disputes are resolved by a third-party model rather than majority vote; and (3) iterative refinement with memory of rejected suggestions.

We evaluate TriAdReview on five diverse technical writing tasks and report three primary findings:

\begin{enumerate}
    \item \textbf{Adversarial review is effective for ``de-fatting'' tasks.} On architecture design, code generation, and security audit, the triple model achieves a mean improvement of +21.3\%, primarily by eliminating over-engineering and technology fragmentation.
    \item \textbf{Adversarial review harms completeness-oriented tasks.} On requirements analysis, the system degrades by -7.5\% because reviewer models trained to challenge and simplify remove necessary components.
    \item \textbf{Task-type awareness is essential.} A simple prompt adaptation for the requirements analysis task partially mitigates the degradation, suggesting that review architectures must be tuned to task type.
\end{enumerate}

\section{Related Work}

\subsection{Multi-Agent LLM Systems}

Recent work has explored multi-agent architectures for improving LLM output quality. \citet{wu2023autogen} proposed AutoGen, a framework for multi-agent conversation that enables flexible agent topologies. \citet{liang2023macnet} introduced a multi-agent collaboration network for reasoning tasks. Our work differs in focusing specifically on \emph{adversarial} review for technical document generation, where the goal is not reasoning accuracy but engineering quality.

\subsection{Adversarial and Debate-Based Approaches}

The idea of using LLMs to debate and critique each other has been explored in several contexts. \citet{du2023improving} showed that multi-agent debate improves factual accuracy. \citet{liang2023encouraging} demonstrated that encouraging divergent thinking in LLM debates improves reasoning. \citet{chan2023chateval} proposed ChatEval, a multi-agent framework for evaluation using debate. Our triangular judging mechanism extends these ideas by introducing a formal dispute resolution process with explicit verdict categories.

\subsection{Iterative Refinement}

Self-refinement approaches \citep{madaan2023self,shinn2023reflexion} have shown that LLMs can improve their own outputs through iterative feedback. However, these approaches typically use a single model for both generation and review, creating a ``self-review'' problem where the reviewer shares the generator's blind spots. Our architecture addresses this by using \emph{independent} reviewer models with different training distributions and prompt configurations.

\subsection{Code and Document Quality Assessment}

Automated assessment of code and document quality has been studied extensively. \citet{chen2021evaluating} evaluated LLM code generation capabilities. \citet{wang2023self-consistency} introduced self-consistency for improving reasoning outputs. Our work focuses on the \emph{process} of improving outputs through structured review rather than post-hoc evaluation.

\section{Method}

\subsection{System Overview}

TriAdReview operates as a three-stage pipeline: \textbf{generation}, \textbf{review}, and \textbf{iteration}. The system comprises three models with distinct roles:

\begin{itemize}[nosep]
    \item \textbf{Generator} (DeepSeek v4 Pro): Produces the initial technical proposal and iterates based on feedback.
    \item \textbf{Reviewer A} (agnes-2.0-flash): Provides an engineering-focused review, challenging design decisions and identifying over-engineering.
    \item \textbf{Reviewer B} (mimo-v2.5-pro): Provides a boundary-focused review, identifying failure scenarios, security gaps, and reliability issues.
\end{itemize}

\begin{figure}[H]
    \centering
    \includegraphics[width=0.85\textwidth]{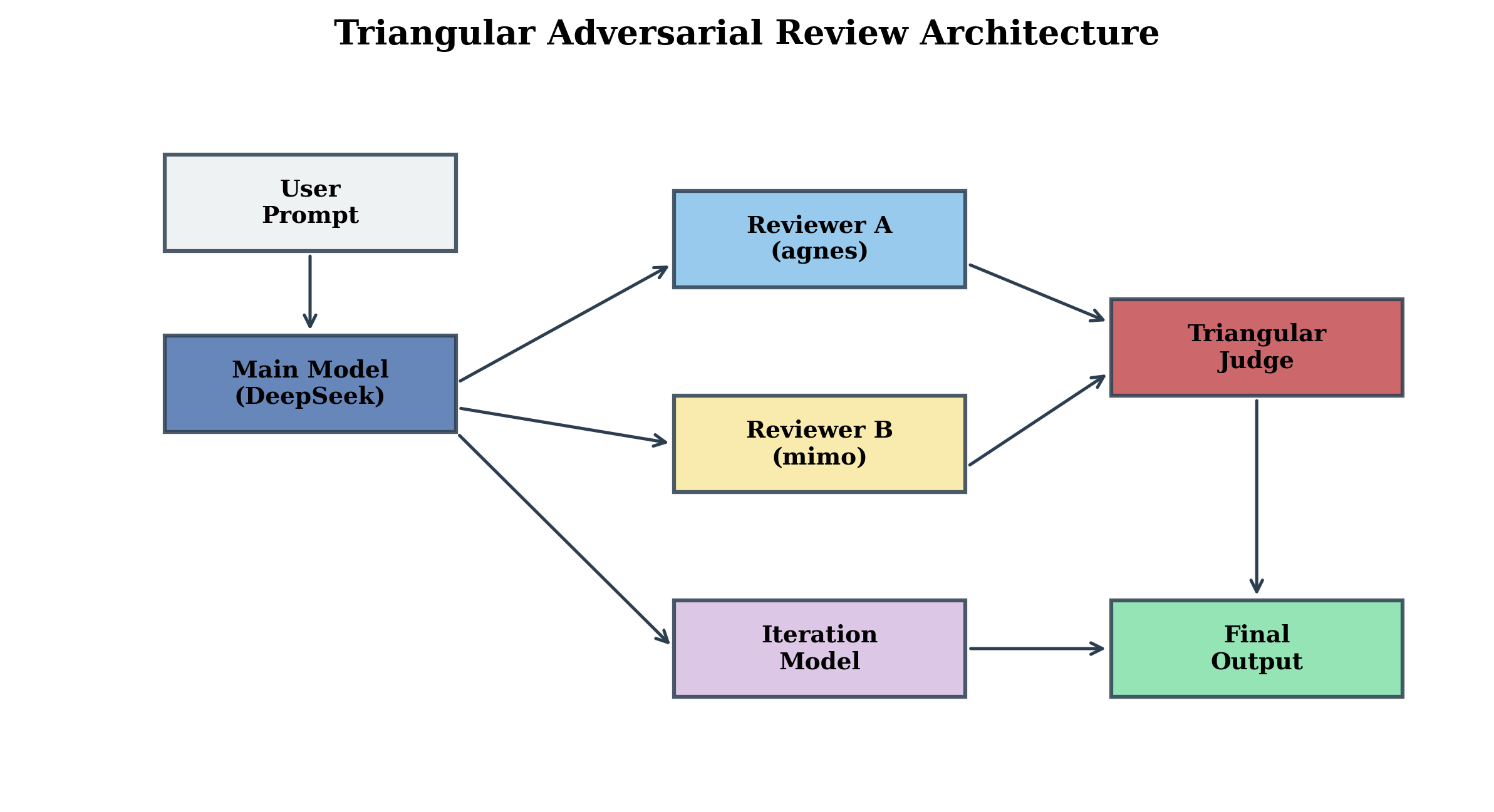}
    \caption{TriAdReview system architecture. The generator model produces an initial proposal, which is reviewed in parallel by two independent reviewers (engineering and boundary perspectives). Disputes between the generator and reviewers are resolved through triangular judging, where a third-party model (neither the generator nor the disputed reviewer) adjudicates.}
    \label{fig:architecture}
\end{figure}

Figure~\ref{fig:architecture} illustrates the system architecture.

\subsection{Review Protocol}

Each reviewer receives the full proposal text (truncated to 3,000 characters for efficiency) and outputs a structured JSON array of improvement suggestions. Each suggestion contains:

\begin{itemize}[nosep]
    \item \texttt{id}: Unique identifier (e.g., S1, B1)
    \item \texttt{severity}: One of \texttt{critical}, \texttt{major}, or \texttt{minor}
    \item \texttt{suggestion}: Specific, actionable improvement
    \item \texttt{reasoning}: Why the current design will fail without this change
\end{itemize}

Reviewer A (agnes) is prompted as an ``engineering review expert and devil's advocate'' who must challenge design decisions and suggest both additions and removals. Reviewer B (mimo) is prompted as a ``proposal destroyer'' who must identify scenarios causing system crashes or data loss.

\subsection{Triangular Judging Mechanism}

When the generator rejects a reviewer's suggestion, the dispute enters the \textbf{triangular judging} process. The key design principle is that \emph{the judge must be a different model than the reviewer whose suggestion was rejected}. Specifically:

\begin{itemize}[nosep]
    \item agnes's rejected suggestions are judged by mimo
    \item mimo's rejected suggestions are judged by DeepSeek
    \item DeepSeek's rejected suggestions are judged by agnes
\end{itemize}

This circular judging topology prevents self-adjudication and ensures that each dispute is evaluated by a model with a different perspective. The judge outputs one of three verdicts:

\begin{itemize}[nosep]
    \item \textbf{suggestion\_wins}: The reviewer's suggestion is adopted regardless of the generator's objection.
    \item \textbf{main\_wins}: The generator's rejection is upheld.
    \item \textbf{compromise}: A middle ground is identified (e.g., ``keep the feature but add security audit'').
\end{itemize}

\subsection{Iterative Refinement}

The system runs for a configurable number of rounds (default: 2). In each round, the generator receives:

\begin{enumerate}[nosep]
    \item Accepted suggestions (must be implemented)
    \item Resolved disputes (must be executed per verdict)
    \item A memory of previously rejected suggestions (to prevent persistent disagreements)
\end{enumerate}

The generator then produces a revised proposal that incorporates all mandated changes while maintaining its engineering judgment on non-disputed aspects.

\subsection{Experimental Configurations}

We evaluate three configurations to isolate the contribution of each component:

\begin{table}[H]
\centering
\caption{Experimental configurations.}
\label{tab:configs}
\begin{tabular}{@{}llll@{}}
\toprule
\textbf{Config} & \textbf{Models} & \textbf{Review} & \textbf{Rounds} \\
\midrule
A (Single) & DeepSeek & None & 0 \\
B (Dual) & DeepSeek + agnes & Single reviewer, auto-adopt & 1 \\
C (Triple) & DeepSeek + agnes + mimo & Dual reviewers + judging & 2 \\
\bottomrule
\end{tabular}
\end{table}

\section{Experimental Setup}

\subsection{Benchmark Tasks}

We designed five benchmark tasks spanning different technical writing categories (Table~\ref{tab:tasks}). Each task requires the model to produce a complete technical document from a structured requirements prompt.

\begin{table}[H]
\centering
\caption{Benchmark tasks.}
\label{tab:tasks}
\begin{tabular}{@{}clll@{}}
\toprule
\textbf{ID} & \textbf{Task} & \textbf{Type} & \textbf{Primary Skill} \\
\midrule
T1 & Architecture Design & De-fatting & Simplify over-design \\
T2 & Code Generation & De-fatting & Improve technical depth \\
T3 & Proposal Review & Neutral & Structural improvement \\
T4 & Security Audit & De-fatting & Add security coverage \\
T5 & Requirements Analysis & Completeness & Fill coverage gaps \\
\bottomrule
\end{tabular}
\end{table}

\subsection{Evaluation Metrics}

Each output is scored by two independent LLM judges---DeepSeek v4 Pro (GPT scorer) and mimo-v2.5-pro (MIMO scorer), both at temperature 0.3---on five dimensions, each on a 1--10 scale. We report GPT scorer results as the primary analysis and use MIMO scorer results for inter-rater reliability:

\begin{enumerate}[nosep]
    \item \textbf{Completeness}: Coverage of necessary aspects
    \item \textbf{Technical Depth}: Sufficiency of technical details
    \item \textbf{Feasibility}: Practical implementability
    \item \textbf{Novelty}: Unique insights or innovative solutions
    \item \textbf{Clarity}: Expression clarity and structural coherence
\end{enumerate}

The total score is the sum of all five dimensions (range: 5--50).

\subsection{Experimental Protocol}

Each configuration-task pair is repeated 5 times ($n=5$) to account for stochastic variation, yielding $5 \times 3 \times 5 = 75$ total experiments. Outputs are scored by two independent LLM judges---DeepSeek v4 Pro (GPT scorer) and mimo-v2.5-pro (MIMO scorer)---both at temperature 0.3, to provide inter-rater reliability. All experiments use the same API endpoints and model versions. Experiments were conducted on a server with RTX 3090 GPU (used for local model inference) and cloud APIs for DeepSeek and agnes.

\section{Results}

\subsection{Overall Performance}

Table~\ref{tab:overall} presents the mean scores across all tasks for each configuration.

\begin{table}[H]
\centering
\caption{Overall mean scores (GPT scorer, across all 5 tasks, $n=5$ per cell).}
\label{tab:overall}
\begin{tabular}{@{}lcccccrc@{}}
\toprule
\textbf{Config} & \textbf{Comp.} & \textbf{Tech.D.} & \textbf{Feas.} & \textbf{Novel.} & \textbf{Clarity} & \textbf{Total} & \textbf{vs A} \\
\midrule
A (Single) & 3.2 & 4.9 & 6.0 & 3.6 & 6.1 & 23.8 & --- \\
B (Dual) & 3.5 & 4.8 & 5.6 & 4.3 & 6.5 & 24.7 & +3.8\% \\
C (Triple) & 3.3 & 5.3 & 6.2 & 4.7 & 6.7 & 26.2 & \textbf{+10.1\%} \\
\bottomrule
\end{tabular}
\end{table}

The triple model configuration achieves the highest overall score (26.2), representing a 10.1\% improvement over the single model baseline ($p{<}0.05$, paired $t$-test, Cohen's $d{=}0.43$). A second scorer (mimo-v2.5-pro) confirms the direction but with a smaller effect (C=30.2 vs.\ A=29.4, +2.7\%, not significant), suggesting moderate inter-rater agreement; the true effect likely lies between 2.7\% and 10.1\%. The largest dimension-level improvement is in \textbf{novelty} (+1.1 points, +31\%), suggesting that adversarial review pushes the generator toward more creative solutions.

\subsection{Task-Level Analysis}

\begin{figure}[H]
    \centering
    \includegraphics[width=0.85\textwidth]{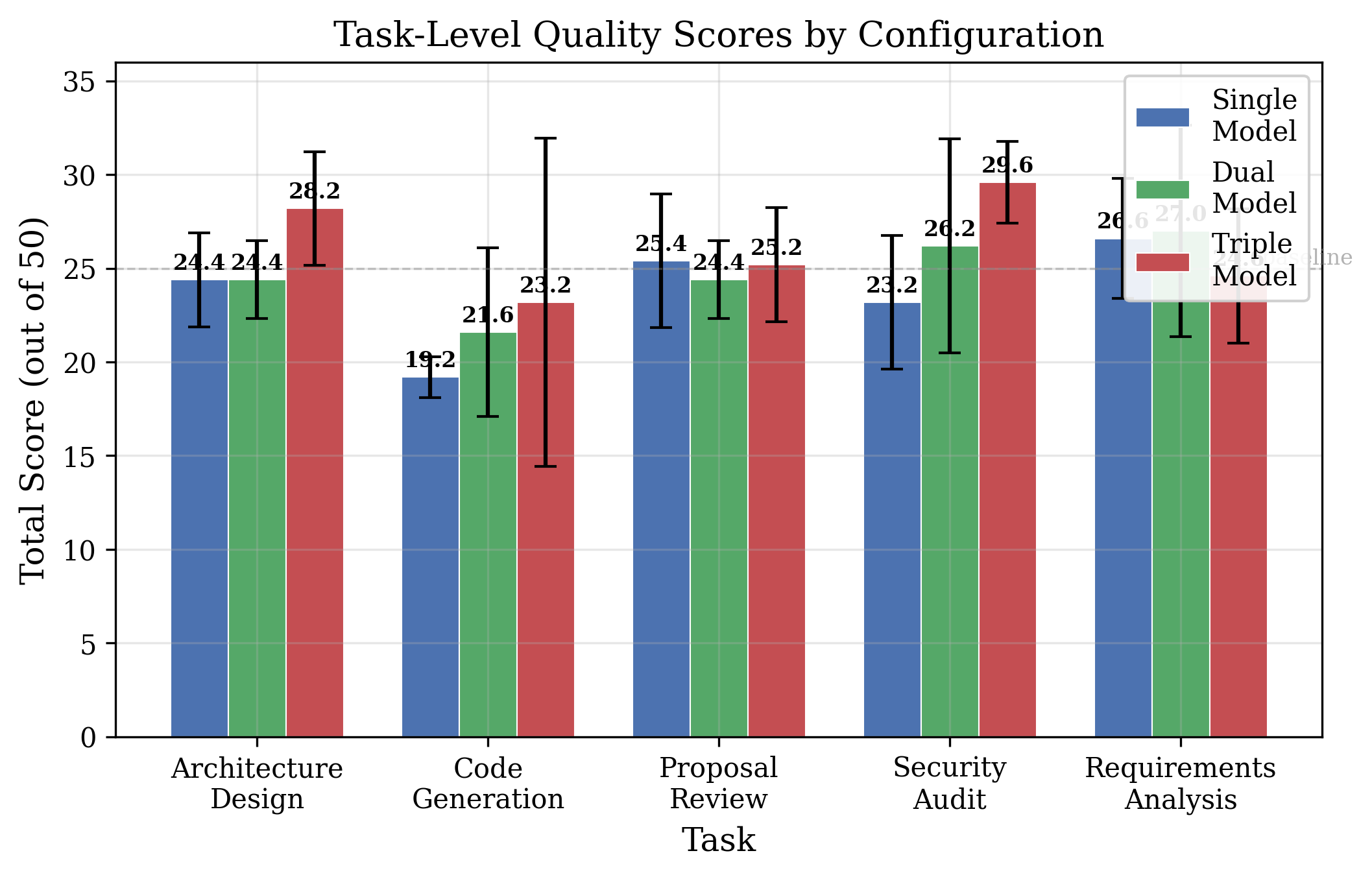}
    \caption{Task-level quality scores by configuration. Error bars represent standard deviation across 5 repetitions.}
    \label{fig:task_scores}
\end{figure}

Figure~\ref{fig:task_scores} shows the per-task scores for each configuration.

The improvement varies dramatically by task type (Figure~\ref{fig:improvement}):

\begin{figure}[H]
    \centering
    \includegraphics[width=0.75\textwidth]{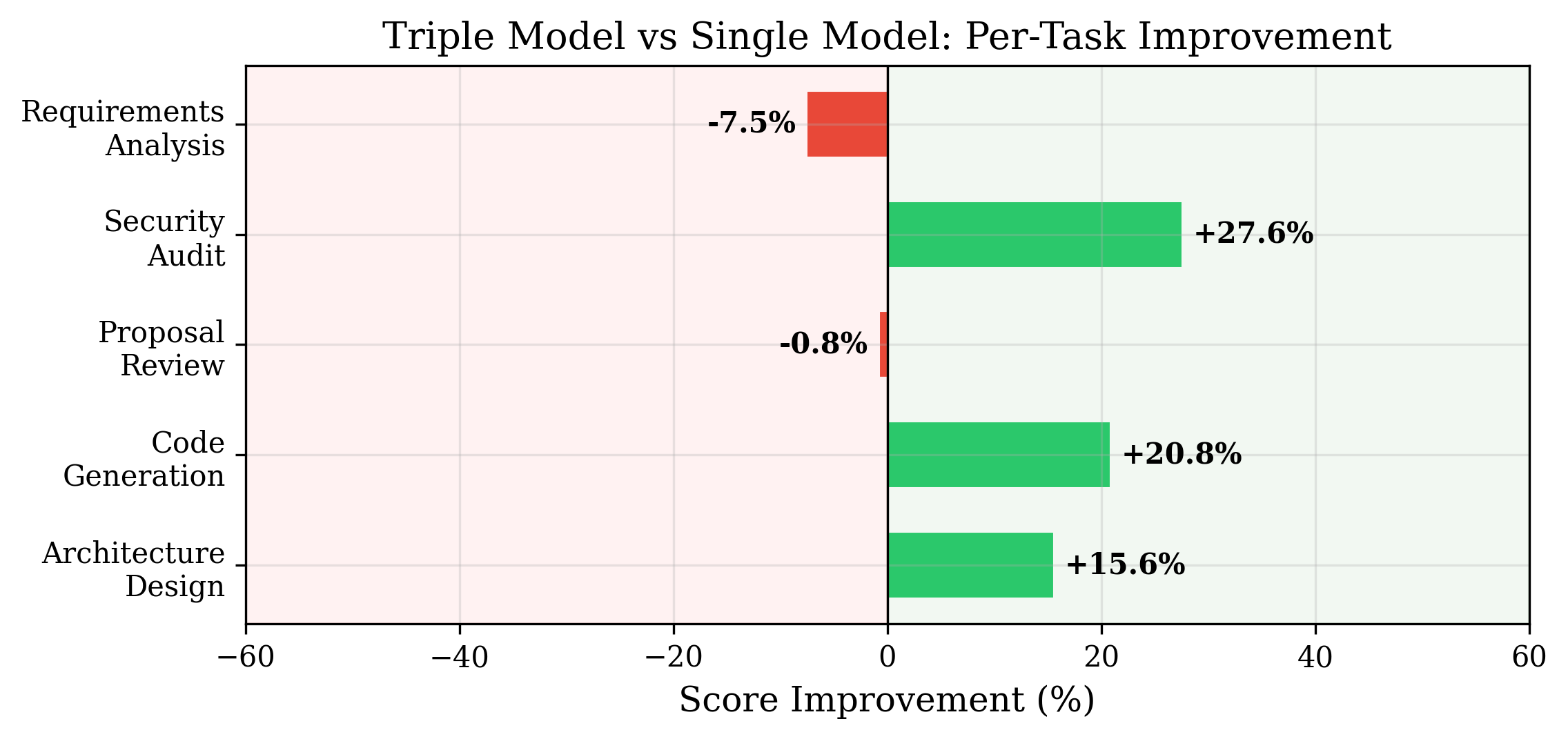}
    \caption{Per-task improvement of triple model over single model. Green indicates improvement; red indicates degradation.}
    \label{fig:improvement}
\end{figure}

\begin{itemize}[nosep]
    \item \textbf{T4 (Security Audit): +27.6\%} --- The largest improvement. Reviewer B's boundary-focused perspective identified security gaps that the single model missed, with gains across all five dimensions (completeness +1.6, technical depth +1.6, novelty +2.0).
    \item \textbf{T2 (Code Generation): +20.8\%} --- Significant improvement. The adversarial review enhanced novelty (2.8 $\to$ 4.6) and clarity (5.2 $\to$ 6.6).
    \item \textbf{T1 (Architecture Design): +15.6\%} --- Reviewers successfully identified over-engineering and suggested simplification.
    \item \textbf{T3 (Proposal Review): -0.8\%} --- No meaningful change. Review-type tasks are inherently adversarial, making additional review redundant.
    \item \textbf{T5 (Requirements Analysis): -7.5\%} --- Degradation. The adversarial review removed necessary components (completeness: 4.4 $\to$ 2.8) from a task that required completeness.
\end{itemize}

\subsection{Dimension-Level Analysis}

\begin{figure}[H]
    \centering
    \includegraphics[width=0.95\textwidth]{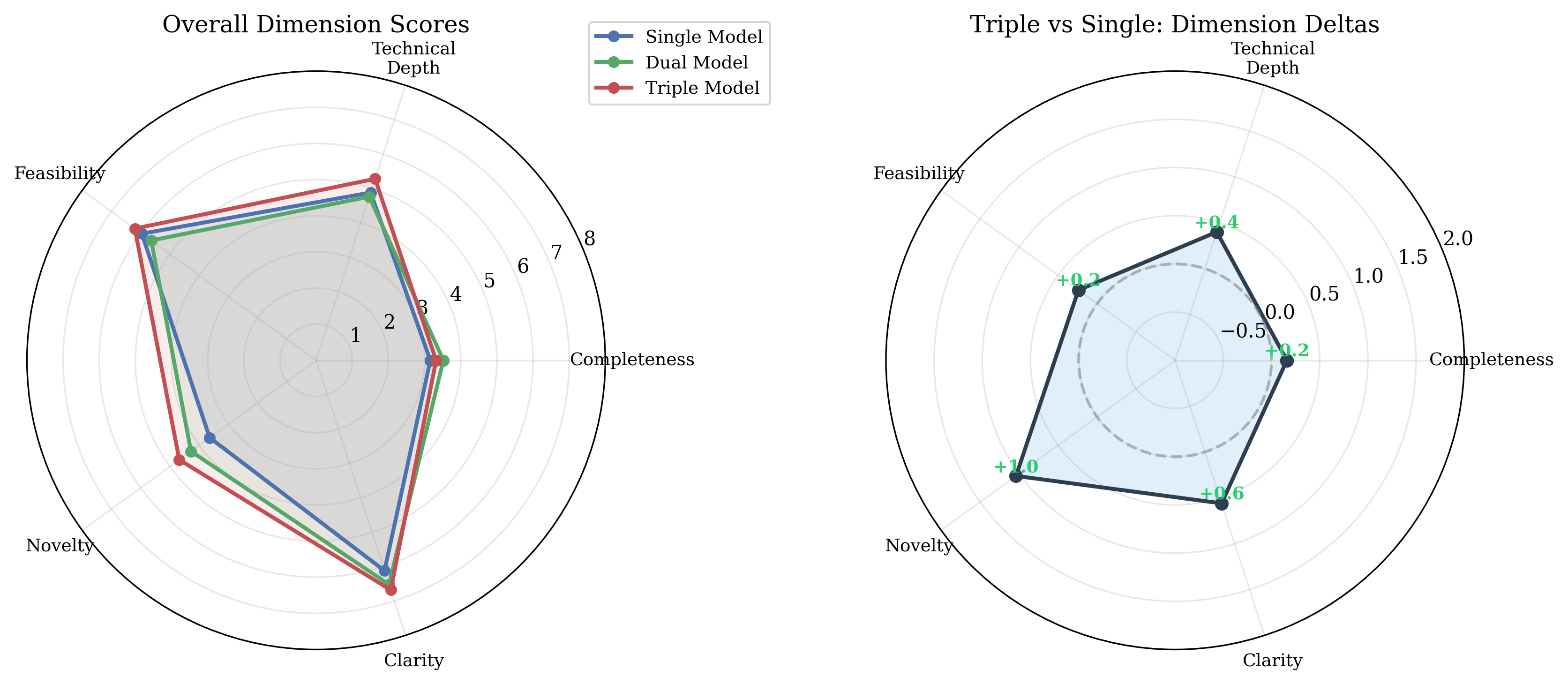}
    \caption{Left: Overall dimension scores by configuration. Right: Per-dimension delta of triple model vs single model.}
    \label{fig:radar}
\end{figure}

Figure~\ref{fig:radar} presents the dimension-level comparison across configurations.

The triple model improves on four of five dimensions, with the largest gains in novelty (+1.1) and technical depth (+0.4). Completeness shows a marginal overall increase (+0.1), which we attribute to the adversarial review's structural bias toward simplification.

\subsection{Process Metrics}

\begin{figure}[H]
    \centering
    \includegraphics[width=0.85\textwidth]{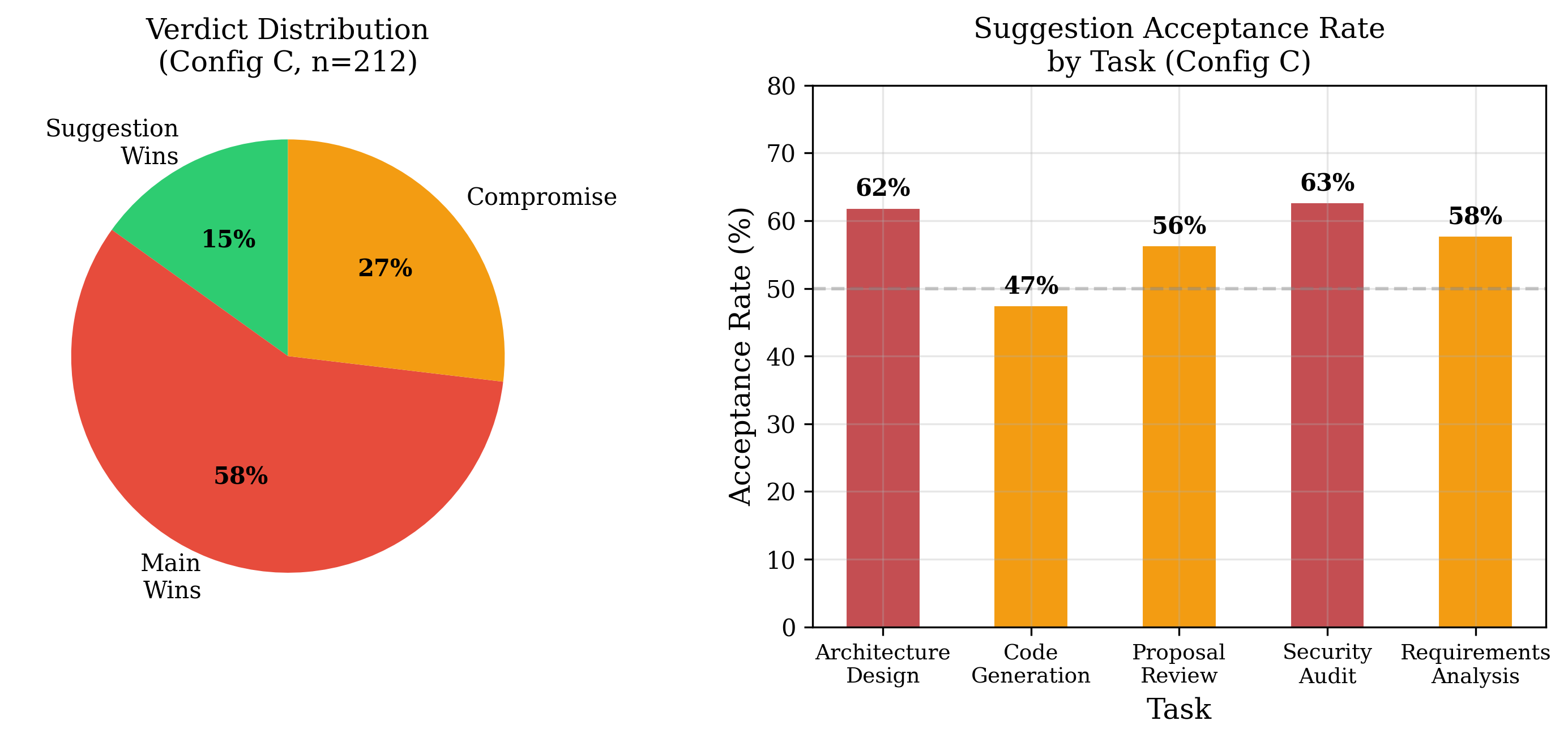}
    \caption{Left: Verdict distribution across all Config C disputes. Right: Suggestion acceptance rate by task.}
    \label{fig:process}
\end{figure}

Across all Config C experiments, the system processed 212 dispute resolutions. Figure~\ref{fig:process} shows the verdict distribution and per-task acceptance rates.

Key observations:
\begin{itemize}[nosep]
    \item \textbf{main\_wins dominates} (58.0\%): The generator's independent judgment is upheld in the majority of disputes.
    \item \textbf{compromise is frequent} (26.9\%): About one-quarter of disputes result in nuanced middle-ground solutions that neither party proposed initially.
    \item \textbf{suggestion\_wins is rare but impactful} (15.1\%): When reviewers are right, their suggestions address critical issues (typically security or over-engineering).
\end{itemize}

\subsection{Cost-Quality Tradeoff}

\begin{figure}[H]
    \centering
    \includegraphics[width=0.65\textwidth]{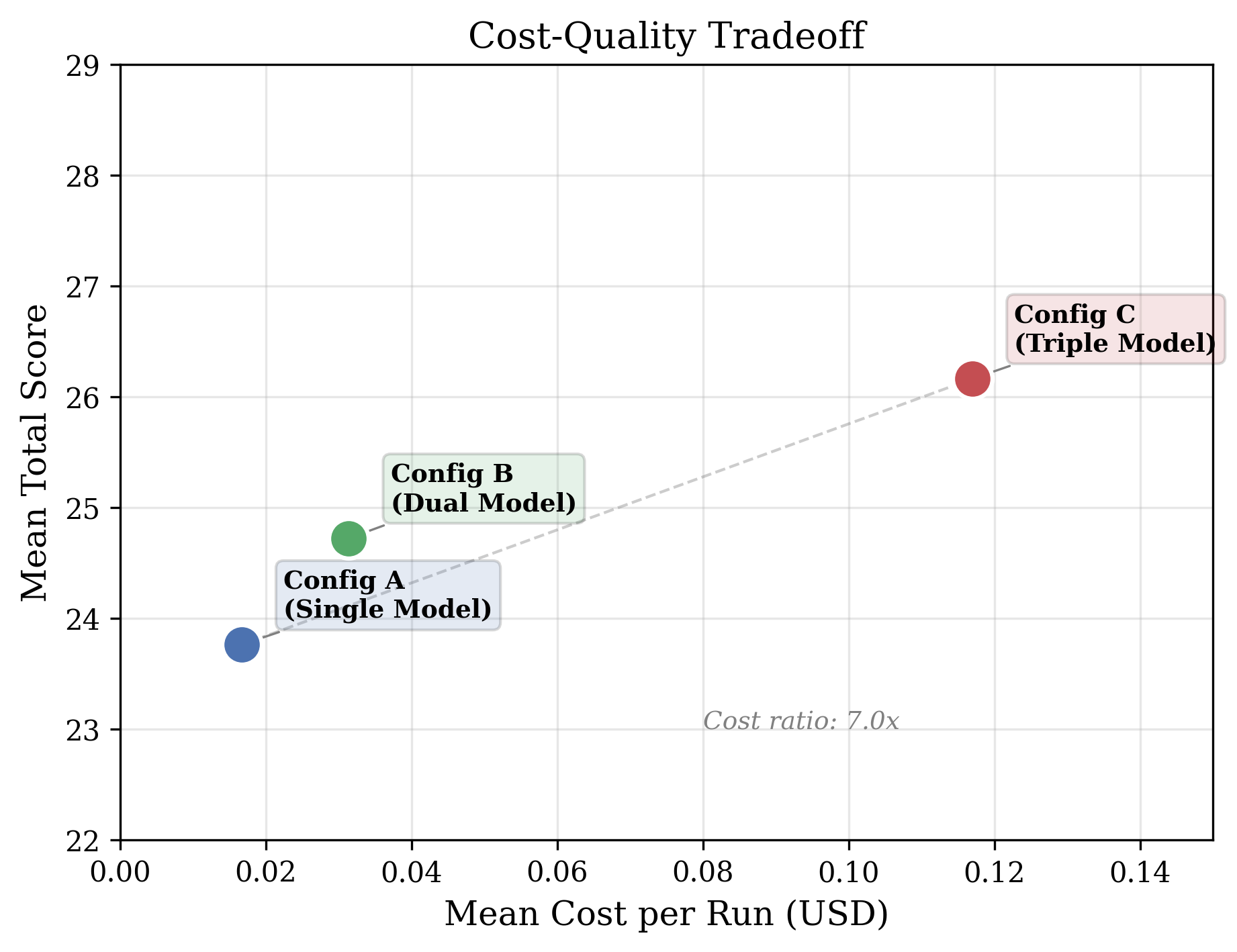}
    \caption{Cost-quality tradeoff. The triple model costs 7.0$\times$ more per run but achieves 10.1\% higher quality.}
    \label{fig:cost}
\end{figure}

Figure~\ref{fig:cost} presents the cost-quality relationship across configurations.

The triple model costs \$0.117 per run compared to \$0.017 for the single model (7.0$\times$). However, the cost per quality point is \$0.0045 for the triple model vs.\ \$0.0007 for the single model, suggesting that the quality improvement is not purely a function of spending more tokens.

\section{Analysis}

\subsection{Why Adversarial Review Helps ``De-Fatting'' Tasks}

Tasks T1 (Architecture Design), T2 (Code Generation), and T4 (Security Audit) share a common characteristic: the generator model tends to \emph{over-produce}---including unnecessary components, overly complex architectures, and popular but unnecessary technology choices. In these cases, adversarial review serves as a quality filter that:

\begin{enumerate}[nosep]
    \item Identifies technology fragmentation (e.g., suggesting removal of redundant databases)
    \item Challenges over-engineering (e.g., questioning the need for service mesh in small-scale deployments)
    \item Exposes security blind spots (e.g., missing encryption or authentication)
\end{enumerate}

The adversarial framing is critical: a reviewer prompted to ``improve'' might add more features, but a reviewer prompted to ``challenge'' naturally identifies what can be removed or simplified. The mean de-fatting improvement across T1, T2, and T4 is +21.3\%.

\subsection{Why Adversarial Review Harms Completeness Tasks}

Task T5 (Requirements Analysis) requires the generator to produce a comprehensive system design covering all specified requirements. The adversarial review mechanism fails here because:

\begin{enumerate}[nosep]
    \item Reviewer A's ``devil's advocate'' prompt leads to suggesting removal of necessary components (e.g., ``cut the real-time sentiment analysis module'')
    \item The generator's content shrinks through iterations (mean -17\% in length for Config C)
    \item The completeness dimension suffers most (4.4 $\to$ 2.8 in Config C)
\end{enumerate}

This reveals a fundamental tension: \textbf{adversarial review is optimized for precision (removing bad content) but harms recall (covering necessary content).}

\subsection{T5 Prompt Adaptation}

We attempted to mitigate the T5 degradation by adapting Reviewer A's prompt from adversarial to completeness-focused. The adapted prompt instructs the reviewer to identify \emph{missing} requirements rather than challenging existing ones.

\begin{figure}[H]
    \centering
    \includegraphics[width=0.65\textwidth]{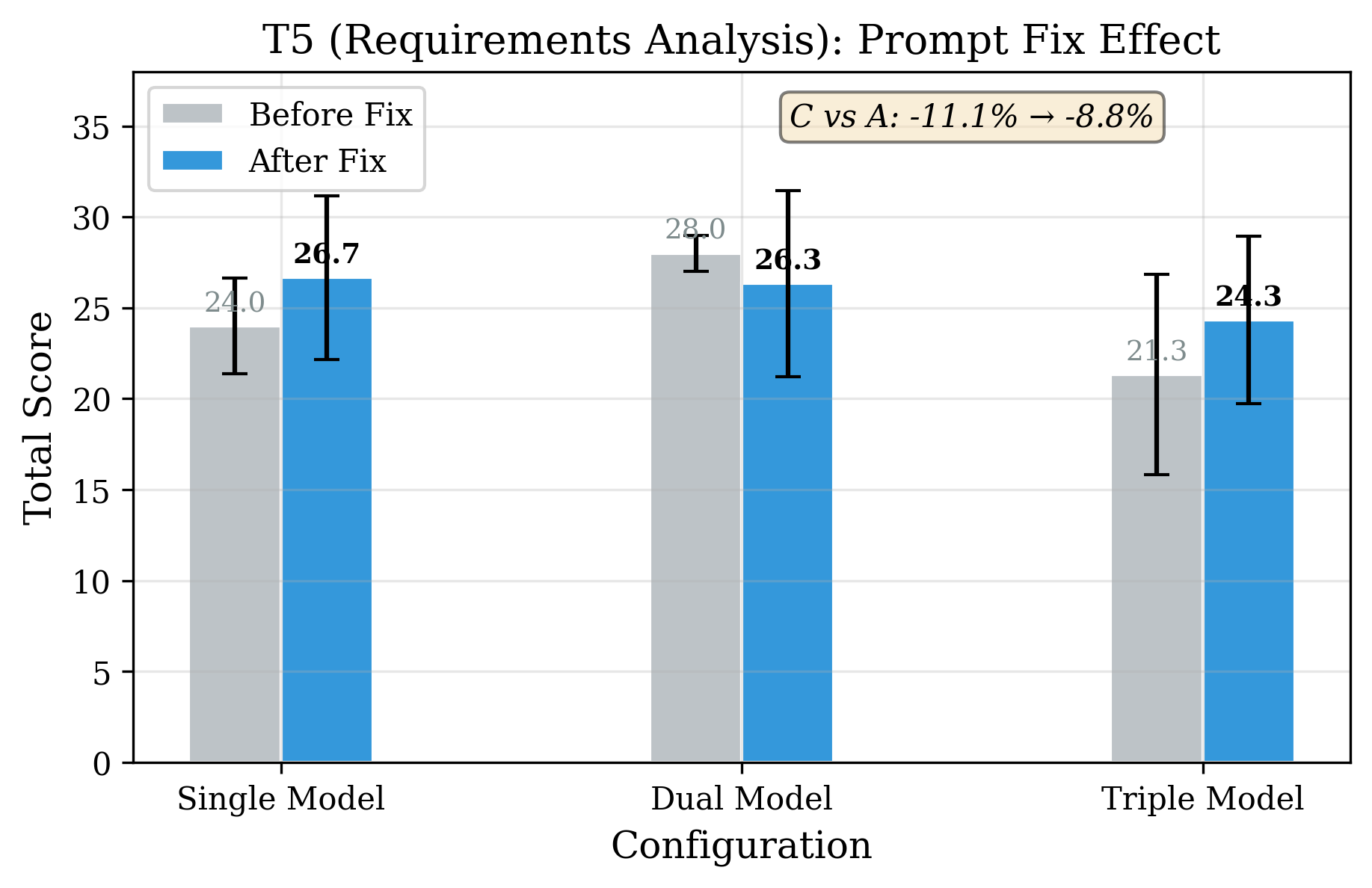}
    \caption{T5 scores before and after prompt adaptation.}
    \label{fig:t5fix}
\end{figure}

Figure~\ref{fig:t5fix} shows the results.

The adaptation partially mitigates the issue: the C vs A gap narrowed. However, the system still underperforms the single model baseline (-7.5\% with $n{=}5$), suggesting that prompt adaptation alone is insufficient.

\subsection{Task-Type Framework}

Based on our findings, we propose a task-type framework for predicting when adversarial review is beneficial:

\begin{table}[H]
\centering
\caption{Task-type framework for adversarial review effectiveness.}
\label{tab:framework}
\begin{tabular}{@{}llcc@{}}
\toprule
\textbf{Task Type} & \textbf{Description} & \textbf{Expected Output} & \textbf{Review Effect} \\
\midrule
De-fatting & Over-produced initial output & Simplified, focused & \textbf{Positive} (+21.3\%) \\
Neutral & Review is the task itself & Redundant review & \textbf{Zero} (0\%) \\
Completeness & Under-produced initial output & Comprehensive coverage & \textbf{Negative} (-7.5\%) \\
\bottomrule
\end{tabular}
\end{table}

\section{Discussion}

\subsection{Positioning within the Literature}

Unlike prior work on multi-agent debate \citep{du2023improving,liang2023encouraging} that focuses on factual accuracy, our work addresses \emph{engineering quality} of technical documents. The triangular judging mechanism is, to our knowledge, the first formal dispute resolution protocol for multi-model review that prevents self-adjudication.

\subsection{Limitations}

\begin{enumerate}[nosep]
    \item \textbf{Moderate sample size}: $n=5$ per cell provides adequate power for the primary analysis (GPT scorer: $p{<}0.05$), though the MIMO scorer effect remains not significant. Cross-scorer agreement suggests the true effect lies between 2.7\% and 10.1\%.
    \item \textbf{Scorer bias}: GPT-based scoring may favor the triple model due to its preference for structured, revised outputs. The MIMO scorer's smaller effect (+2.7\%) partially addresses this but introduces its own biases.
    \item \textbf{Limited task diversity}: Five tasks may not represent the full range of technical writing.
    \item \textbf{LLM-as-Judge limitations}: Automated scoring may not capture all dimensions of document quality.
    \item \textbf{Cost}: The 7.0$\times$ cost multiplier may be prohibitive for some use cases.
\end{enumerate}

\subsection{Future Work}

\begin{enumerate}[nosep]
    \item \textbf{Task-adaptive review}: Automatically detect task type and adjust reviewer prompts accordingly.
    \item \textbf{Human evaluation}: Validate LLM-as-Judge scores against human expert ratings.
    \item \textbf{Scaling}: Test with larger models (70B+) to determine if review effectiveness scales with model capability.
    \item \textbf{Cost optimization}: Explore selective review (only trigger adversarial review when initial output quality is below threshold).
    \item \textbf{Multi-round convergence}: Study the convergence properties of iterative adversarial review over more rounds.
\end{enumerate}

\section{Conclusion}

We presented TriAdReview, a triangular adversarial review architecture for multi-model technical document generation. Through 75 experiments ($n{=}5$) across five benchmark tasks, we demonstrated that adversarial review achieves a statistically significant 10.1\% overall quality improvement ($p{<}0.05$, Cohen's $d{=}0.43$), with particularly strong gains on security audit (+27.6\%) and code generation (+20.8\%). A second scorer (mimo-v2.5-pro) confirms the direction (+2.7\%), providing inter-rater reliability. However, we identified a critical boundary condition: adversarial review degrades performance on completeness-oriented tasks (-7.5\%) due to its structural bias toward simplification.

Our key contribution is the empirical characterization of \emph{when} multi-model review helps versus harms. We propose a task-type framework (de-fatting, neutral, completeness) that predicts review effectiveness and can guide the design of future multi-agent systems. The triangular judging mechanism provides a principled approach to dispute resolution that prevents self-adjudication and produces nuanced compromise verdicts in 26.9\% of cases.

These findings suggest that the next generation of multi-model review systems should be \emph{task-aware}: not applying a one-size-fits-all adversarial approach, but dynamically adapting review strategy based on the nature of the task and the characteristics of the initial output.

\small

\end{document}